\documentclass{article}

    \PassOptionsToPackage{numbers, compress}{natbib}
\usepackage[preprint]{main}

\usepackage[utf8]{inputenc} % allow utf-8 input
\usepackage[T1]{fontenc}    % use 8-bit T1 fonts
\usepackage{hyperref}       % hyperlinks
\usepackage{url}            % simple URL typesetting
\usepackage{booktabs}       % professional-quality tables
\usepackage{amsfonts}       % blackboard math symbols
\usepackage{multirow}       % multirow cells in tables
\usepackage{graphicx}       % for including graphics
\usepackage{nicefrac}       % compact symbols for 1/2, etc.
\usepackage{microtype}      % microtypography
\usepackage{xcolor}         % colors
\usepackage{booktabs}
\usepackage{amssymb} % for ✓ and ✗
\usepackage{booktabs}
\usepackage{pifont}
\usepackage{graphicx}
\usepackage{wrapfig}
\usepackage{graphicx}

\usepackage[frozencache,cachedir=minted-cache]{minted}
\usepackage{float}       % generic floats
\floatstyle{plain}
\newfloat{listing}{t}{lol}   % defines the 'listing' environment
\usepackage{amsmath} 
\usepackage{tabularx}
\newcolumntype{L}[1]{>{\raggedright\arraybackslash}p{#1}} % 左对齐定宽

\title{CodeAgents: A Token-Efficient Framework for Codified Multi-Agent Reasoning in LLMs}

\author{
Bruce Yang$^{1}$,
Xinfeng He$^{2}$,
Huan Gao$^{3}$,
Yifan Cao$^{3}$,
Xiaofan Li$^{1}$,
David Hsu$^{1}$\\
\\
$^1$National University of Singapore\\
$^2$School of Computer Science and Engineering, Southeast University \\
$^3$Sapiens AI \\
\\
\texttt{e1100032@u.nus.edu} \\
\texttt{hexinfeng2003@gmail.com} \\
\texttt{gh@seu.edu.cn} \\
\texttt{pete@sapiens-ai.io} \\
\texttt{li.x@nus.edu.sg} \\
\texttt{dcsdavid@nus.edu.sg}
}

\begin{document}

\maketitle

\begin{abstract}

Effective prompt design is essential for improving the planning capabilities of large language model (LLM)-driven agents. However, existing structured prompting strategies are typically limited to single-agent, plan-only settings, and often evaluate performance solely based on task accuracy—overlooking critical factors such as token efficiency, modularity, and scalability in multi-agent environments. To address these limitations, we introduce CodeAgents, a prompting framework that codifies multi-agent reasoning enables structured, token-efficient planning in multi-agent systems. In CodeAgents, all components of agent interaction—\texttt{Task}, \texttt{Plan}, \texttt{Feedback}, system roles, and external tool invocations—are codified into modular pseudocode enriched with control structures (e.g., loops, conditionals), boolean logic, and typed variables. This design transforms loosely connected agent plans into cohesive, interpretable, and verifiable multi-agent reasoning programs.We evaluate the proposed framework across three diverse benchmarks—\textbf{GAIA}, \textbf{HotpotQA}, and \textbf{VirtualHome}—using a range of representative LLMs. Results show consistent improvements in planning performance, with absolute gains of 3–36 percentage points over natural language prompting baselines. On VirtualHome, our method achieves a new state-of-the-art success rate of 56\%. In addition, our approach reduces input and output token usage by 55–87\% and 41–70\%, respectively, underscoring the importance of token-aware evaluation metrics in the development of scalable multi-agent LLM systems. The code and resources are available at: \url{https://anonymous.4open.science/r/CodifyingAgent-5A86}

\end{abstract}

\section{Introduction}
Large Language Models (LLMs) have emerged as powerful planning engines for AI agents, capable of transforming an initial world state into a desired goal state. From web navigation to robotic manipulation, accurate planning is \emph{critical} in LLM-driven agent tasks: a flawed plan can derail the entire reasoning process. Indeed, planning is fundamental to intelligent behavior, and recent work has increasingly focused on equipping LLM agents with robust planning abilities to tackle complex, multi-step tasks. The challenge is that these tasks demand not only correct reasoning but also efficient use of the model’s limited context and resources. This makes it imperative to develop prompting strategies that yield \emph{both} high plan accuracy and token-efficient reasoning.

A growing line of prompting frameworks has begun to address complex reasoning. Techniques such as Chain-of-Thought (CoT) prompting \citep{wei2022cot}, Program-Aided Language Models (PAL) \citep{gao2023pal}, ReAct \citep{yao2022react}, and Tree-of-Thoughts \citep{yao2023treeofthoughts} enable LLMs to decompose problems into intermediate steps or integrate tool use. However, most existing methods follow a \textbf{single-agent} paradigm: one monolithic LLM handles planning and reasoning in a continuous, verbose natural language stream. This design leads to several limitations. First, reasoning traces tend to be overly verbose, causing high token usage and latency. Recent studies show that standard CoT solutions often use significantly more tokens than necessary, highlighting significant inefficiency \citep{wen2025codeplan}. Second, conflating all roles (planner, solver, verifier) into one sequence can hinder clarity and correctness. LLMs often stumble when simultaneously planning and executing solutions, as errors can be compounded without clear modular structures \citep{singh2023progprompt}. Finally, current prompting strategies largely ignore the \emph{cost} of tokens—they aim to maximize task success but do not explicitly minimize token consumption. This oversight is problematic as context window limits and API costs become bottlenecks for deploying LLM agents. In summary, today’s prompting approaches for complex tasks are not token-efficient, are limited to single-agent scenarios, and rely on free-form language that may be unnecessarily verbose or ambiguous.

To address these issues, we propose CodeAgents: a framework for token-efficient multi-agent reasoning that encodes multi-agent problem solving as structured pseudocode. The key idea is to \textbf{treat a complex reasoning task like a program}, with different modules (or agents) handling planning, execution, and feedback in a coordinated way. Instead of prompting the model with a verbose natural language dialogue, we provide a pseudocode template that the LLM fills in and follows. This pseudocode explicitly represents \textbf{multi-agent interactions}: a \texttt{Planner} outlines high-level plans, a \texttt{Solver} executes detailed reasoning, and a \texttt{Reviewer} provides feedback. Components communicate clearly through well-defined variables, iterating as needed within a coherent prompt. Our structured approach significantly reduces ambiguity and verbosity, prompting the model in a concise, algorithmic style.A summary comparison of Codifying-Agent against existing prompting strategies is shown in Table~\ref{tab:prompting-comparison}, where it demonstrates higher success rates and lower token costs while supporting dynamic replanning. 

Crucially, our framework introduces several technical innovations to enhance expressivity and efficiency. First, we incorporate \textbf{typed variables}, enabling clear distinctions between different data types and reducing errors. Second, we leverage \textbf{control flow} structures such as loops and conditionals, enabling dynamic reasoning strategies and branching logic, unlike linear natural-language prompts. Third, we create \textbf{reusable subroutines} encapsulating common reasoning patterns, improving modularity and token efficiency. Finally, our entire prompting approach prioritizes \textbf{token-cost awareness}, explicitly optimizing prompt length and token usage, making efficient use of LLM resources without sacrificing reasoning quality.

We validate our framework on challenging benchmarks, demonstrating its effectiveness. Our contributions can be summarized as follows:

\begin{itemize}
\item \textbf{Code-First Multi-Agent Prompting:} We introduce a novel framework encoding complex reasoning tasks as modular pseudocode, orchestrating specialized agents (planning, reasoning, feedback) within a single LLM, significantly reducing verbosity.

\item \textbf{Token Efficiency as a Core Metric:} We explicitly optimize token usage in prompt design and evaluation, achieving superior efficiency and practicality under strict context limits.

\item \textbf{Empirical Performance Gains:} Extensive experiments demonstrate that our framework consistently outperforms existing prompting strategies on benchmarks including GAIA\citep{mialon2023gaia}, HotpotQA\citep{yang2018hotpotqa}, and VirtualHome\citep{puig2018virtualhome}, delivering higher accuracy and solution quality with substantially reduced token usage.
\end{itemize}

\begin{table}[t]
\centering
\caption{Comparison of prompting strategies for LLM planning and execution. \textbf{Codifying-Agent} achieves better success with fewer tokens via structured re-usable plans.}
\label{tab:prompting-comparison}
\begin{tabular}{lccc}
\toprule
\textbf{Method} & \textbf{Token Cost (M)} & \textbf{Success Rate} & \textbf{Supports Replanning} \\
\midrule
Chain-of-Thought \citep{wei2022cot} & High & Moderate & \ding{55}\\
Plan-and-Solve \citep{wang2023planandsolve} & High & Moderate & \ding{55} \\
ProgPrompt \citep{singh2023progprompt} & Low & Medium & \ding{55} \\
ReAct \citep{yao2022react} & High & Medium & \checkmark \\
\textbf{CodeAgents} & \textbf{Low} & \textbf{High} & \textbf{\checkmark} \\
\bottomrule
\end{tabular}
\end{table}

%%%%%%%%%%%%%%%%%%%%%%%%%%%%%%%%%%%%%%%%%%%%%%%%%%%%%%%%%%%%%%%%%%%%%%%%%%%%
%  Section 2 – Methodology 
%%%%%%%%%%%%%%%%%%%%%%%%%%%%%%%%%%%%%%%%%%%%%%%%%%%%%%%%%%%%%%%%%%%%%%%%%%%%
\section{Methodology}
\label{sec:methodology}

We propose a Codified Prompting framework that reformulates all components of LLM-driven agent reasoning into a unified representation of structured, typed pseudocode. This includes agent roles, task decomposition plans, tool invocations, intermediate feedback, and observational outcomes. By encoding these elements programmatically, the framework enhances interpretability, reduces token overhead, and improves the traceability of multi-step reasoning processes.
Unlike conventional prompting methods that rely on a single agent emitting unconstrained natural language, our approach orchestrates a multi-agent system composed of specialized roles—such as \texttt{Planner}, \texttt{ToolCaller}, and \texttt{Replanner}. Each agent is tasked with a specific functional responsibility, and their interactions are mediated through modular pseudocode messages. These messages incorporate typed variables, conditional logic, and control structures, thereby supporting composability and reusability of reasoning steps.

By shifting from natural language to a code-like intermediate representation, the framework supports precise coordination, error localization, and automated evaluation of reasoning traces. This code-first design also facilitates downstream integration with external tools and APIs, enabling more robust agent behavior in complex environments. Overall, Codified Prompting enables structured, interpretable, and cost-efficient LLM-based reasoning through a principled abstraction over multi-agent interaction.

\begin{figure}[t]
  \centering
  \includegraphics[width=\linewidth]{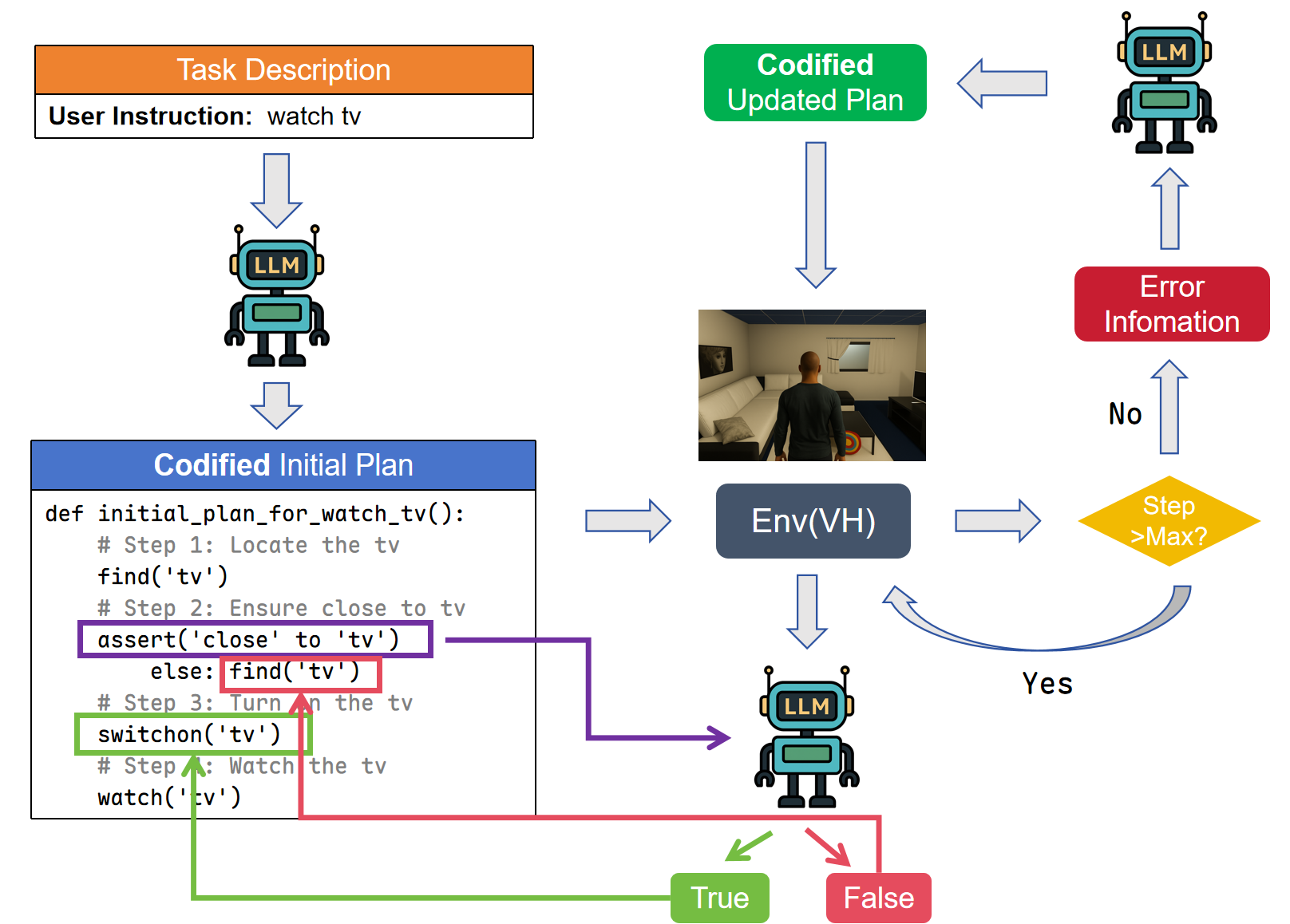}
  \caption{Architecture of our VirtualHome agent loop. The agent executes modular plans containing in-plan \texttt{assert} checks for local error recovery and invokes a global feedback loop to re-plan on critical failures.}
  \label{fig:virtualhome-arch}
\end{figure}

\subsection{Single-Agent Codified Reasoning Framework}
\label{sec:virtualhome}

Figure~\ref{fig:virtualhome-arch} illustrates our single-agent architecture, demonstrated on the \textit{watch TV} task in a simulated VirtualHome environment. The agent generates structured pseudocode plans and executes them step-by-step, using runtime feedback for iterative replanning. Assertion checks are embedded to catch and locally recover from errors, or escalate to global plan revisions when needed. A concrete case is shown in Figure~\ref{fig:eat-bread}.

\textbf{Structured Task Prompting.}
The system prompt defines available actions and objects using Python-like syntax. Typed variables and few-shot examples guide the LLM to generate semantically aligned pseudocode, concluding with a partial function header to elicit structured completions.

\textbf{Function-Based Planning.}
Each task is encoded as a named function composed of sequential action calls with typed arguments. Inline comments clarify intent and context, enhancing plan interpretability. While not executable, the code serves as a formal, readable scaffold for reasoning.

\textbf{Precondition Assertions.}
Assertions check critical preconditions (e.g., object visibility, possession) before actions. Failures invoke local recovery routines (e.g., \texttt{else} branches) to avoid unnecessary replanning and ensure resilience to minor errors.

\textbf{Feedback-Driven Replanning.}
Major failures trigger a replanning loop: the system constructs a structured diagnostic trace—including the failed step, error type, and current state—which is passed back to the LLM for sub-plan regeneration. This loop enables robustness in dynamic, long-horizon tasks.

\begin{figure}[ht]
\centering
\begin{minipage}[t]{0.5\textwidth}
\noindent\textbf{Task Prompt:}
\begin{minted}[fontsize=\small, bgcolor=gray!5]{python}
from actions import walk<obj>, grab<obj>, 
    find<obj>, sit<obj>, eat<obj>
objects = ["sofa", "bread", ...]

# Example tasks
def throw_away_apple():
    ...
def watch_tv():
    ...

# Next Task
def initial_plan_for_eat_bread_on_sofa():

\end{minted}

\noindent\textbf{LLM Response:}
\begin{minted}[fontsize=\small, bgcolor=gray!5]{python}
def initial_plan_for_eat_bread_on_sofa():
    # Step 1: Locate sofa and bread
    walk('livingroom')
    find('sofa')
    # Step 2: Pick up the bread
    assert('close' to 'bread')
        else: find('bread')
    grab('bread')
    # Step 3: Sit on the sofa
    sit('sofa')
    # Step 4: Eat the bread
    eat('bread')
\end{minted}
\end{minipage}
\hfill
\begin{minipage}[t]{0.48\textwidth}
\noindent\textbf{Error Feedback:}
\begin{minted}[fontsize=\small, bgcolor=yellow!5]{python}
def initial_plan_for_eat_bread_on_sofa():
    # Step 1: Locate sofa and bread
    ...
    # Step 2: Pick up the bread
    ...
    # Step 3: Sit on the sofa
    ...
error_step = "sit('sofa')"
feedback_message = (
  "not close to <sofa> when [SIT]"
)
environmental_information = [
  "sofa is visible but not near",
  "agent is holding bread"
]
items_in_hand = ["bread"]
def updated_plan_for_eat_bread_on_sofa():

\end{minted}

\noindent\textbf{LLM Response:}
\begin{minted}[fontsize=\small, bgcolor=yellow!5]{python}
def updated_plan_for_eat_bread_on_sofa():
    # Step 3: Ensure near the sofa
    assert('close' to 'sofa')
        else: find('sofa')
    sit('sofa')
    # Step 4: Eat the bread
    eat('bread')
\end{minted}
\end{minipage}
\caption{Example of a fully codified prompt for the task \texttt{eat\_bread\_on\_sofa}, combining code, comments, assertions, and replanning.}
\label{fig:eat-bread}
\end{figure}

\begin{figure}
    \centering
    \includegraphics[width=1.0\linewidth]{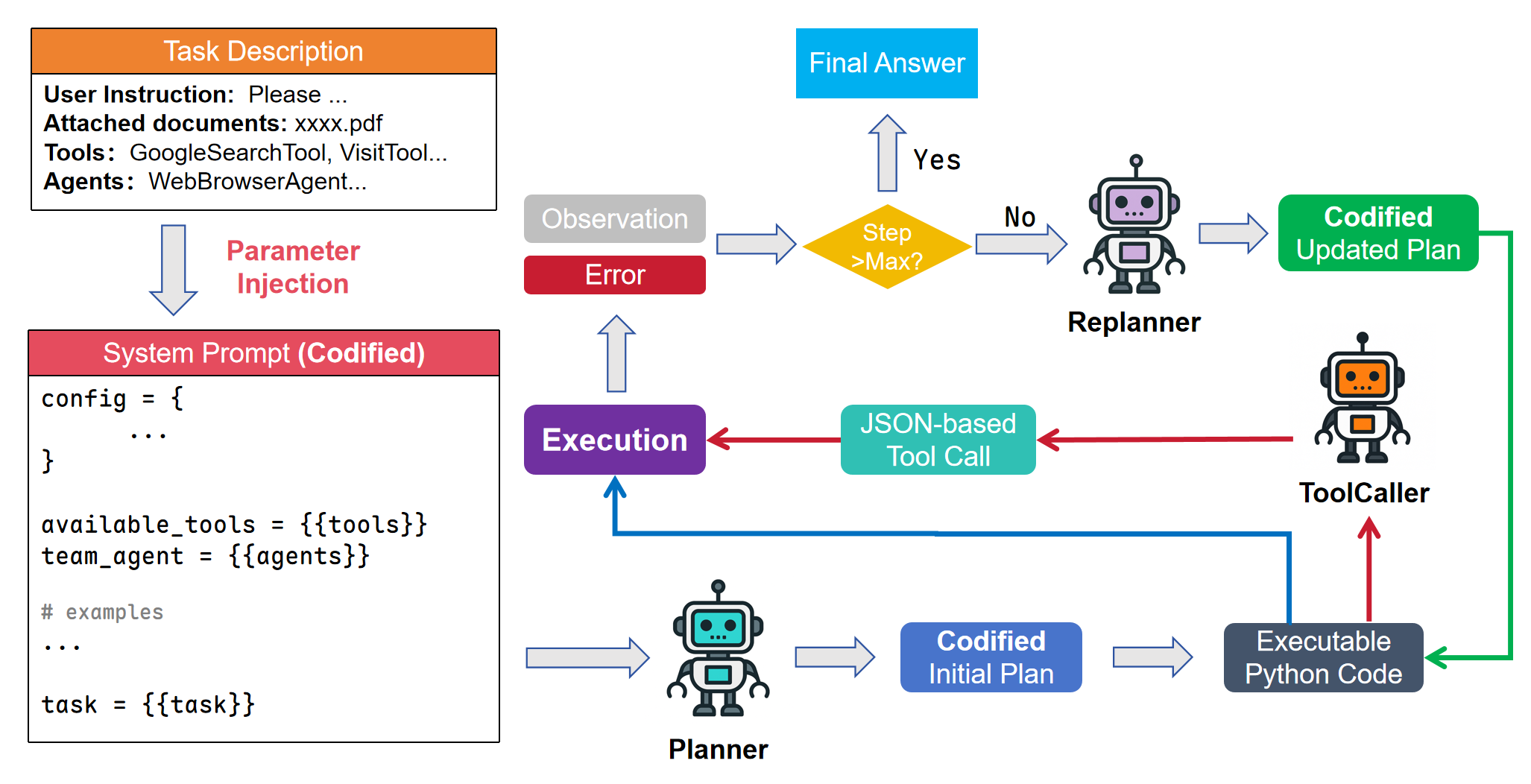}
    \caption{CodeAgents architecture for multi-agent coordination. Agents exchange codified prompts, plans, and executable tool calls to coordinate planning, execution, and replanning in a modular loop.}
    \label{fig:multi-agent-arch}
\end{figure}

\subsection{Multi-Agent Codified Reasoning Framework}
\label{sec:gaia-hotpot}

The CodeAgents framework includes both single-agent and multi-agent variants, which share a unified code-first prompting paradigm using structured pseudocode. These variants differ in architectural configuration: the single-agent version integrates planning, execution, and feedback into one loop, while the multi-agent version distributes these roles across specialized agents. Though methodologically consistent, the two designs are optimized for different task characteristics and are not interchangeable. We present both to illustrate the generality and adaptability of our approach across diverse reasoning scenarios.

Figure~\ref{fig:multi-agent-arch} illustrates the architecture of the proposed \textit{Multi-Agent Codified Reasoning Framework}, wherein specialized agents—namely \texttt{Planner}, \texttt{ToolCaller}, and \texttt{Replanner}—collaborate through structured code-based exchanges to solve complex reasoning tasks. The workflow is initiated by a codified system prompt derived from user instructions and contextual resources. Agents communicate via pseudocode plans, executable tool invocations, and structured feedback, thereby forming a modular and iterative reasoning loop. This architecture facilitates transparent agent coordination, modular control over execution, and systematic error recovery. A representative interaction trace is shown in Figure~\ref{fig:multi-agent-example}.

\textbf{Codified System Prompt.} Each agent is initialized using a structured, codified system prompt in YAML format. The prompt specifies the agent’s role (e.g., \texttt{expert\_assistant}), available tools, teammate agents, and execution cycle (e.g., \texttt{[thought, code, observation]}). This declarative format enables precise capability control, modular prompt reuse, and interoperability between agents via code-oriented input-output schemas.

\textbf{Structured Plan Generation.}
Upon receiving a user-defined instruction, the \texttt{Planner} agent synthesizes a high-level plan encoded as Python-style pseudocode. The plan consists of declarative steps involving variable instantiations, conditional logic, and iterative structures. Each action is annotated with natural language comments, which serve as interpretable intermediate thoughts to guide the LLM’s decision-making. Compared to traditional natural language Chain-of-Thought prompting, this codified planning paradigm reduces semantic ambiguity, supports plan verification, and facilitates seamless handoff to downstream agents for execution or revision.

% \textbf{Executable \texttt{CodeAct} and Tool Invocation.}
% The \texttt{ToolCaller} agent transforms abstract pseudocode steps into concrete API calls, referred to as \texttt{CodeAct} instructions—an approach inspired by prior work on code-based action representations~\citep{wang2024codeact}. These calls correspond to executable functions (e.g., web search, environment interaction) and are typically serialized in structured formats such as JSON or executable Python. Unlike monolithic single-agent systems, our framework decouples tool invocation from planning, enabling targeted validation and asynchronous execution. Upon invocation, results—whether success signals or retrieved content—are codified and returned for use in subsequent reasoning stages.【修正错误，2025.05.20】

\textbf{Executable \texttt{CodeAct} and Tool Invocation.}
The \texttt{Planner} produces an initial plan in the form of structured pseudocode, which is subsequently transformed—also by the planner—into executable \texttt{CodeAct}, written in a Python syntax. This codified representation captures both logical flow and concrete tool interaction instructions. Once generated, the \texttt{CodeAct} can follow two execution paths: it can either be parsed by the \texttt{ToolCaller} to perform structured tool invocations (e.g., JSON-based API calls), or it can be executed directly via a Python interpreter to perform general computations. This flexible execution strategy decouples planning from execution, supports targeted validation, and enables iterative re-planning in response to runtime feedback or errors.

\textbf{Code-Based Feedback and Replanning.}
When tool execution fails or produces unsatisfactory observations, the \texttt{Replanner} agent is activated. It consumes structured error traces encoded as pseudocode, which include the failed instruction, execution context, and associated metadata (e.g., agent ID, timestamp, tool state). Based on this structured diagnostic input, the \texttt{Replanner} synthesizes a revised sub-plan that resumes execution from the point of failure. This agent-level modular replanning mechanism enhances system robustness, supports adaptive recovery, and enables generalization to dynamic or partially observable environments.

\begin{figure}[ht]
\centering
\begin{minipage}[t]{0.49\textwidth}
\vspace{-0.3em}\textbf{System Prompt:}\vspace{-0.3em}
\begin{minted}[fontsize=\small, bgcolor=gray!5, breaklines]{python}
config = {
  role: 'expert_assistant',
  tools: 'python_callable',
  cycle: [thought, code, observation]
}
available_tools = [
  GoogleSearchTool, VisitTool, PageUpTool,
  PageDownTool, FinderTool, TextInspectorTool
]
team_agent = [web_browser_agent]
task = "Search the web and summarize 3 key growth drivers in AI chip market"
\end{minted}
\vspace{-0.3em}\textbf{Initial Plan:}\vspace{-0.3em}
\begin{minted}[fontsize=\small, bgcolor=gray!5, breaklines]{python}
def initial_plan():
  # Step 1: Use Google to find relevant article
  query = "AI chip market 2025 site:techcrunch.com"
  search_result = GoogleSearchTool(query)
  doc = search_result[0]['url']
  # Step 2: Open the link
  VisitTool(doc)
  # Step 3: Scroll until key term appears
  while not TextInspectorTool.contains("growth driver"):
    PageDownTool()
    if too_many_pages_scrolled: break
  # Step 4: Extract and summarize
  paragraphs = FinderTool(keyword="AI chip")
  drivers = TextInspectorTool(text=paragraphs, 
    focus="growth drivers", count=3)
  final_answer(drivers)
\end{minted}
\end{minipage}
\hfill
\begin{minipage}[t]{0.49\textwidth}
\vspace{-0.3em}\textbf{Executable Python Code:}\vspace{-0.3em}
\begin{minted}[fontsize=\small, bgcolor=yellow!5, breaklines]{python}
url = GoogleSearchTool("AI chip growth driver")[0]['url']
VisitTool(url)
PageDownTool()
text = FinderTool("AI chip")
drivers = TextInspectorTool(text=text, focus="growth drivers", count=3)
\end{minted}
\vspace{-0.3em}\textbf{Tool Call Examples:}\vspace{-0.3em}

\begin{minted}[fontsize=\small, bgcolor=yellow!5]{json}
{
  "tool": "GoogleSearchTool",
  "args": {
    "query": "AI chip market 2025
    site:techcrunch.com"
  }
}
\end{minted}

\vspace{-0.3em}\textbf{Error Feedback:}\vspace{-0.4em}
\begin{minted}[fontsize=\small, bgcolor=yellow!5]{python}
error_message = "URL failed to load"
failed_url = 
"https://techcrunch.com/example-ai-chip"
\end{minted}

\vspace{-0.3em}\textbf{Updated Plan:}\vspace{-0.4em}
\begin{minted}[fontsize=\small, bgcolor=yellow!5, breaklines]{python}
def updated_plan():
  # Step 2: Retry different URL
  url = GoogleSearchTool("AI chip summary")[0]['url']
  VisitTool(url)
  # Step 3: Scroll
  PageDownTool()
  # Step 4: Extract and summarize
  text = FinderTool("AI chip")
  drivers = TextInspectorTool(text=text, 
    focus="growth drivers", count=3)
\end{minted}

\vspace{-0.3em}\textbf{Output Observation:}\vspace{-0.3em}
\begin{minted}[fontsize=\small, bgcolor=yellow!5]{python}
driver1 = "Edge computing boom"
driver2 = "National policy support"
driver3 = "Generative AI demand"
\end{minted}
\end{minipage}
\caption{A simplified codified agent workflow illustrating structured system prompt initialization, pseudocode planning, executable tool invocation, and feedback-driven replanning.}
\label{fig:multi-agent-example}
\end{figure}

%%%%%%%%%%%%%%%%%%%%%%%%%%%%%%%%%%%%%%%%%%%%%%%%%%%%%%%%%%%%%%%%%%%%%%%%%%%%
% End of Section 2
%%%%%%%%%%%%%%%%%%%%%%%%%%%%%%%%%%%%%%%%%%%%%%%%%%%%%%%%%%%%%%%%%%%%%%%%%%%%

\section{Experiment}

This section presents a comprehensive evaluation of code-based prompting strategies across three representative benchmarks: VirtualHome, GAIA, and HotpotQA. The goal of these experiments is to rigorously assess the impact of structured code-style instructions on the performance and efficiency of LLM agents. The section is structured to elaborate on the experimental setup, employed models, evaluation metrics, and detailed analysis of results in each benchmark setting.

Our methodology is evaluated on two tracks:
\begin{itemize}
    \item \textbf{VirtualHome\citep{puig2018virtualhome}:} A high-fidelity, single-agent benchmark for evaluating long-horizon task planning in dynamic environments. Our approach draws inspiration from ProgPrompt~\citep{singh2023progprompt}, which introduced code-form planning with local \texttt{assert}-based error handling, and BrainBody~\citep{bhat2024brainbody}, which employed natural language plans augmented with feedback and replanning mechanisms. While both achieved strong results on VirtualHome, they differ in how they handle failures—local vs. global, code vs. language. We use this benchmark to systematically test various combinations of code, natural language, and hybrid formats for planning and recovery, aiming to identify the most effective tradeoff between task accuracy and token efficiency.
    
    \item \textbf{GAIA\citep{mialon2023gaia} and HotpotQA\citep{yang2018hotpotqa}:} We evaluate our framework on two multi-agent reasoning benchmarks that demand advanced coordination, tool usage, and multi-hop reasoning. \textbf{GAIA} (General AI Assistant) is a recently proposed benchmark designed to assess AI assistants' capabilities in real-world scenarios requiring reasoning, web browsing, and tool use. It comprises 466 human-annotated questions across various domains, emphasizing tasks that are straightforward for humans but challenging for AI systems. GAIA evaluates an agent's proficiency in integrating multiple skills to provide accurate answers. \textbf{HotpotQA} is a large-scale question-answering dataset with 113K multi-hop questions that require reasoning across multiple Wikipedia documents, challenging models to provide both accurate and interpretable answers—making it a valuable benchmark for evaluating multi-agent coordination and reasoning capabilities.
\end{itemize}

\subsection{Evaluation on the VirtualHome Benchmark}

VirtualHome is a 3D simulation platform designed to assess agent behavior in household environments. Tasks involve interpreting natural language instructions, generating executable plans, and interacting with virtual objects to complete daily routines.
We follow the evaluation protocol from ProgPrompt~\cite{singh2023progprompt}, utilizing the same train-validation-test split. Our configuration was run five times using Gemini-2.0-Flash, and we report the mean and standard deviation. We focus on two key metrics: (1) \textbf{Partial Success Rate (PSR)}, which captures the degree of goal completion even if not perfectly executed, and (2) \textbf{Success Rate (SR)}, which measures the proportion of tasks fully completed without error.

\begin{table}[htbp]
\centering
\small
\caption{Comparison of Formats on VirtualHome. For baselines \textit{ProgPrompt} and \textit{BrainBody}, token statistics were not reported in their original papers. We additionally reproduced a natural language variant of our method following the BrainBody scheme.}
\label{tab:virtualhome-full}
\resizebox{\linewidth}{!}{%
\begin{tabular}{lcccccc}
\toprule
\textbf{Format} & \textbf{SR} & \textbf{PSR} & \textbf{Input Token} & \textbf{Output Token} & \textbf{Total Token} \\
\midrule
BrainBody~\cite{bhat2024brainbody} & $0.54 \pm 0.09$ & $0.69 \pm 0.04$ & -- & -- & -- \\
ProgPrompt~\cite{singh2023progprompt} & $0.40 \pm 0.11$ & $0.72 \pm 0.09$ & -- & -- & -- \\
NL(BrainBody) & $0.36 \pm 0.05$ & $0.71 \pm 0.03$ & $10462.02 \pm 730.12$ & $217.58 \pm 20.08$ & $10679.60 \pm 748.21$ \\
\textbf{CodeAgents} & $\mathbf{0.56 \pm 0.11}$ & $\mathbf{0.73 \pm 0.07}$ & $\mathbf{5845.28 \pm 374.69}$ & $\mathbf{484.18 \pm 39.08}$ & $\mathbf{6329.46 \pm 398.07}$ \\
\bottomrule
\end{tabular}
}
\end{table}

Table~\ref{tab:virtualhome-full} compares our approach against prior baselines. Since previous work did not report token consumption, we reproduce a natural language version of our prompt using BrainBody’s schema. Our structured prompting method achieves the highest SR ($\mathbf{0.56}$) and PSR ($\mathbf{0.73}$), while also significantly reducing total tokens (↓40.7\%) compared to the natural language variant. This demonstrates the effectiveness of our codified format in both execution fidelity and efficiency.

\subsection{Evaluation on the GAIA Benchmark}

We evaluate our framework on Level 1 tasks from the GAIA validation set, a benchmark targeting multi-step reasoning, tool usage, and web interaction. Following our methodology, we compare a baseline \textit{natural language format}—where prompts and plans are expressed in free-form text—with our \textit{codified format}. The natural language baseline is constructed using the default prompt template from the SmolAgents framework, which features role-based task descriptions and step-by-step reasoning in conversational form. It is not designed to reflect state-of-the-art performance, but rather to serve as a representative and controlled configuration for isolating the effects of prompt structure. In contrast to SOTA natural language systems that often rely on extensive engineering, expert-crafted prompts, or system-level optimization, our baseline remains untuned and minimal to ensure fairness in comparison. Structured prompts consistently outperform the natural language baseline across models. For Gemini-2.5-Flash, our codified approach improves accuracy by 10.7\%, while reducing input tokens by 67.8\% and cost by 67.4\%. Gemini-2.5-Pro also achieves a 4.8\% accuracy gain with over 40\% cost savings. We attribute these improvements to the high semantic density and reduced ambiguity of the codified format. By eliminating redundant language and standardizing reasoning flows, agents require fewer tokens to complete tool-based reasoning cycles (Table~\ref{tab:GAIA-full}).

\begin{table}[htbp]
\centering
\small
\setlength{\tabcolsep}{6pt}
\renewcommand{\arraystretch}{1.35}
\caption{Comparison of Prompt Formats on Accuracy, Token Usage, and Cost on GAIA}
\begin{tabular}{lccccc}
\toprule
\textbf{Model / Format} & \textbf{Accuracy} & \textbf{Input (M)} & \textbf{Output (K)} & \textbf{Cost (\$)} \\
\midrule
\textbf{Gemini-2.5-Flash} & & & & \\
\quad Natural Language & 0.56 & 72.42 & 314.82 & 11.05 \\
\quad \textbf{CodeAgents} & \textbf{0.62} & \textbf{23.28} & \textbf{174.60} & \textbf{3.60} \\
\quad \textit{Improvement} & \textcolor{red}{+10.7\%} & \textcolor{red}{$\downarrow$67.8\%} & \textcolor{red}{$\downarrow$44.5\%} & \textcolor{red}{$\downarrow$67.4\%} \\
\midrule
\textbf{Gemini-2.5-Pro} & & & & \\
\quad Natural Language & 0.62 & 31.80 & 597.25 & 45.72 \\
\quad \textbf{CodeAgents} & \textbf{0.65} & \textbf{17.93} & \textbf{386.88} & \textbf{26.28} \\
\quad \textit{Improvement} & \textcolor{red}{+4.8\%} & \textcolor{red}{$\downarrow$43.6\%} & \textcolor{red}{$\downarrow$35.2\%} & \textcolor{red}{$\downarrow$42.5\%} \\
\bottomrule
\label{tab:GAIA-full}
\end{tabular}
\end{table}

\subsection{Evaluation on the HotpotQA Benchmark}

\begin{wrapfigure}{r}{0.50\textwidth}
    \centering
    \vspace{-2em}
    \includegraphics[width=\linewidth]{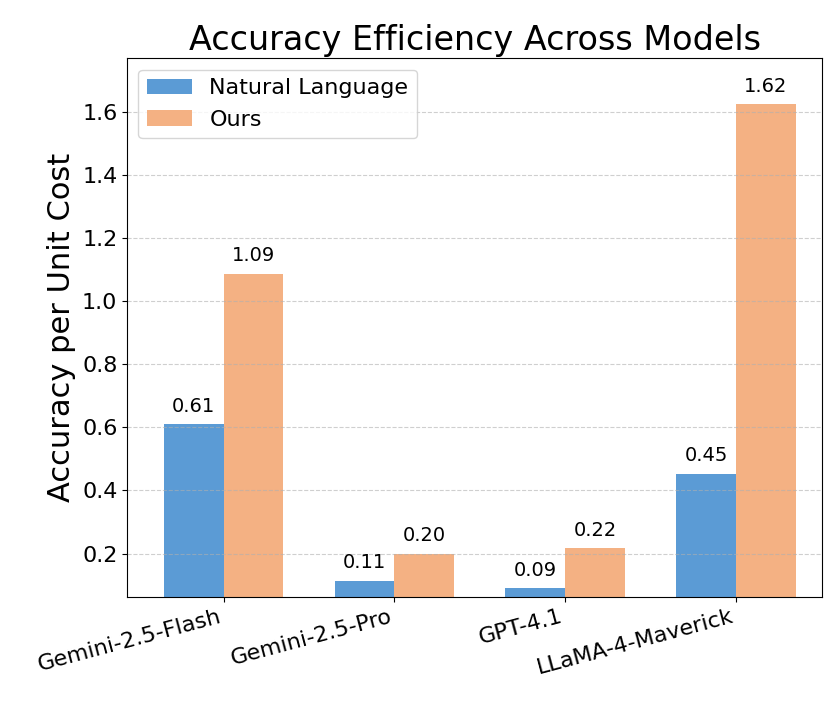}
    \vspace{-2em}
    \caption{Accuracy per unit cost on HotpotQA.}
    \label{fig:hotpotqa-efficiency}
\end{wrapfigure}

We evaluate our framework on HotpotQA, a widely used multi-hop QA benchmark requiring multi-document reasoning. A 100-example subset was sampled from the \texttt{test-fullwiki} split to compare models under natural language and codified prompt formats.
As shown in Table~\ref{tab:hotpotqa-results}, our formats consistently match or outperform natural language in both Accuracy and F1, while significantly reducing token usage and cost. For example, LLaMA-4-Maverick achieves the highest accuracy (0.52) with a 70.4\% reduction in cost. Gemini-2.5-Pro improves average F1 from 0.66 to 0.68 while reducing input tokens by over 50\%.
Figure~\ref{fig:hotpotqa-efficiency} further visualizes \textit{accuracy efficiency} across models. We yield higher performance per unit cost across all models, especially in LLaMA and Gemini-Flash, indicating better semantic density and reduced redundancy.

\vspace{1em}
\begin{table}[htbp]
\centering
\small
\setlength{\tabcolsep}{6pt}
\renewcommand{\arraystretch}{1.35}
\caption{Comparison of Formats on HotpotQA (Accuracy, F1, Token Usage, and Cost)}
\label{tab:hotpotqa-results}
\begin{tabular}{lccccc}
\toprule
\textbf{Model / Format} & \textbf{Accuracy} & \textbf{F1} & \textbf{Input (M)} & \textbf{Output (K)} & \textbf{Cost (\$)} \\
\midrule
\textbf{Gemini-2.5-Flash} & & & & & \\
\quad Natural Language & 0.47 & 0.61 & 4.50 & 164.60 & 0.77 \\
\quad \textbf{CodeAgents} & \textbf{0.50} & \textbf{0.64} & \textbf{2.69} & \textbf{96.66} & \textbf{0.46} \\
\quad \textit{Improvement} & \textcolor{red}{+6.4\%} & \textcolor{red}{+4.9\%} & \textcolor{red}{$\downarrow$40.2\%} & \textcolor{red}{$\downarrow$41.3\%} & \textcolor{red}{$\downarrow$40.3\%} \\
\midrule
\textbf{Gemini-2.5-Pro} & & & & & \\
\quad Natural Language & 0.51 & 0.66 & 2.45 & 139.80 & 4.46 \\
\quad \textbf{CodeAgents} & \textbf{0.51} & \textbf{0.68} & \textbf{1.20} & \textbf{105.50} & \textbf{2.56} \\
\quad \textit{Improvement} & – & \textcolor{red}{+3.0\%} & \textcolor{red}{$\downarrow$51.0\%} & \textcolor{red}{$\downarrow$24.5\%} & \textcolor{red}{$\downarrow$42.6\%} \\
\midrule
\textbf{GPT-4.1} & & & & & \\
\quad Natural Language & 0.46 & 0.63 & 2.11 & 110.74 & 5.10 \\
\quad \textbf{CodeAgents} & \textbf{0.51} & \textbf{0.64} & \textbf{1.00} & \textbf{42.21} & \textbf{2.34} \\
\quad \textit{Improvement} & \textcolor{red}{+10.9\%} & \textcolor{red}{+1.6\%} & \textcolor{red}{$\downarrow$52.6\%} & \textcolor{red}{$\downarrow$61.9\%} & \textcolor{red}{$\downarrow$54.1\%} \\
\midrule
\textbf{LLaMA-4-Maverick} & & & & & \\
\quad Natural Language & 0.49 & 0.63 & 5.71 & 181.63 & 1.08 \\
\quad \textbf{CodeAgents} & \textbf{0.52} & \textbf{0.65} & \textbf{1.58} & \textbf{83.53} & \textbf{0.32} \\
\quad \textit{Improvement} & \textcolor{red}{+6.1\%} & \textcolor{red}{+3.2\%} & \textcolor{red}{$\downarrow$72.3\%} & \textcolor{red}{$\downarrow$54.0\%} & \textcolor{red}{$\downarrow$70.4\%} \\
\bottomrule
\end{tabular}
\end{table}
%%%%%%%%%%%%%%%%%%%%%%%%%%%%%%%%%%%%%%%%%%%%%%%%%%%%%%%%%%%%%%%%%%%%%%%%%%%%
% End of Section 3
%%%%%%%%%%%%%%%%%%%%%%%%%%%%%%%%%%%%%%%%%%%%%%%%%%%%%%%%%%%%%%%%%%%%%%%%%%%%

\section{Ablation Study}

\subsection{VirtualHome Results}
Table~\ref{tab:ablation-VH} reports Success Rate (SR), Partial Success Rate (PSR), Execution Accuracy, and Token usage on VirtualHome. The natural language baseline (Row 1) yields 0.20 SR with high token cost (8280 tokens). Adding \texttt{replan} improves SR to 0.36 but increases tokens by 29\% (Row 2).
Code-only prompting (Row 3) cuts token usage by 88\% while raising SR by 50\%. Adding \texttt{replan} or \texttt{assert} (Rows 4–5) brings further gains, and their combination (Row 6) achieves 0.46 SR and 0.75 execution accuracy—more than double the baseline SR with 42\% fewer tokens.
The best performance (Row 11) comes from combining code, \texttt{assert}, \texttt{replan}, and English comments, reaching 0.56 SR (+180\%) with 24\% fewer tokens. These results highlight the benefits of structured prompting with embedded feedback and lightweight annotations for efficient, accurate reasoning.

\subsection{HotpotQA Results}
Table~\ref{tab:hotpotqa-ablation} summarizes ablation results on HotpotQA. The natural language baseline without \texttt{replan} (Row 1) achieves 41\% accuracy and consumes 4.14 million tokens. Adding \texttt{replan} (Row 2) increases accuracy by 19.5\% but incurs a 42\% token overhead.
Code-based prompting with \texttt{replan} (Row 3) achieves the highest efficiency, improving accuracy by 24.4\% while reducing token usage by 72.2\%. Adding English comments with or without \texttt{replan} (Rows 4–5) further improves interpretability. The best result (Row 5) achieves 52\% accuracy (+26.8\%) with a 60\% reduction in token usage. Including Chinese comments (Row 6) shows comparable accuracy with slightly improved efficiency. The GAIA ablation study shows similar trends and will be available on our GitHub.

\subsection{Analysis}
Across both benchmarks, code-based prompting consistently outperforms natural language baselines in reasoning accuracy and token efficiency. Feedback mechanisms such as \texttt{replan} and \texttt{assert} provide incremental gains by improving execution robustness. Comparing Chinese and English comments, we observe a trade-off: Chinese comments slightly reduce token usage, while English comments yield marginally higher accuracy. These findings support the effectiveness of our modular design and demonstrate the value of integrating structured feedback and annotated code in enhancing LLM reasoning under constrained token budgets.

\begin{table}[htbp]
\centering
\small
\caption{Ablation Study of our methodology on VirtualHome. All improvements ($\Delta$) are calculated relative to the Natural Language baseline (1).}
\label{tab:ablation-VH}
\resizebox{\linewidth}{!}{%
\begin{tabular}{lcccccc}
\toprule
\textbf{Format} & \textbf{SR} & \textbf{PSR} & \textbf{Exec} & \textbf{Total Token} & \textbf{$\Delta$ SR} & \textbf{$\Delta$ Token} \\
\midrule
Natural Language (1) & $0.20$ & $0.62$ & $0.61$ & $8280.00$ & -- & -- \\
Natural Language + Replan (2) & $0.36$ & $0.71$ & $0.63$ & $10679.60$ & {$+80.0\%$} & {$+29.0\%$} \\
\midrule
Code Only (3) & $0.30$ & $0.62$ & $0.68$ & $\mathbf{1010.74}$ & {$+50.0\%$} & {$\downarrow87.8\%$} \\
\quad + Replan (4) & $0.38$ & $0.67$ & $0.70$ & $2624.52$ & {$+90.0\%$} & {$\downarrow68.3\%$} \\
\quad + Assert (5) & $0.36$ & $0.69$ & $0.67$ & $4080.76$ & {$+80.0\%$} & {$\downarrow50.7\%$} \\
\quad + Assert + Replan (6) & $0.46$ & $0.70$ & $\mathbf{0.75}$ & $4767.68$ & {$+130.0\%$} & {$\downarrow42.4\%$} \\
\midrule
\multicolumn{7}{l}{\textit{Code with Comments}} \\
Code + CN Comment + Assert + Replan (7) & $0.52$ & $0.72$ & $0.65$ & $6192.42$ & {$+160.0\%$} & {$\downarrow25.2\%$} \\
Code + EN Comment (8) & $0.34$ & $0.67$ & $0.70$ & $1316.70$ & {$+70.0\%$} & {$\downarrow84.1\%$} \\
\quad + Replan (9) & $0.44$ & $0.72$ & $0.72$ & $3531.58$ & {$+120.0\%$} & {$\downarrow57.3\%$} \\
\quad + Assert (10) & $0.52$ & $0.73$ & $0.69$ & $4137.78$ & {$+160.0\%$} & {$\downarrow50.0\%$} \\
\quad + Assert + Replan (11) & $\mathbf{0.56}$ & $\mathbf{0.73}$ & $0.70$ & $6329.46$ & {$+180.0\%$} & {$\downarrow23.6\%$} \\
\bottomrule
\end{tabular}
}
\end{table}

\begin{table}[htbp]
\centering
\small
\caption{Ablation Study on HotpotQA. All improvements ($\Delta$) are relative to the Natural Language + No Replan baseline (1).}
\label{tab:hotpotqa-ablation}
\begin{tabularx}{\linewidth}{L{5.2cm} cccc cc}
\toprule
\textbf{Format} & \textbf{Acc} & \textbf{F1} & \textbf{Total Token (M)} & $\Delta$Acc & $\Delta$Token \\
\midrule
Natural Language (1) & $0.41$ & $0.53$ & $4.14$ & -- & -- \\
Natural Language + Replan (2) & $0.49$ & $0.63$ & $5.89$ & {$+19.5\%$} & {$\uparrow 42.0\%$} \\
\midrule
Code Only + Replan (3) & $0.51$ & $0.62$ & $\mathbf{1.15}$ & {$+24.4\%$} & {$\downarrow 72.2\%$} \\
\midrule
\textit{Code with Comments} \\
\quad + EN Comment + No Replan (4) & $0.42$ & $0.53$ & $1.19$ & {$+2.4\%$} & {$\downarrow 71.2\%$} \\
\quad + EN Comment + Replan (5) & $\mathbf{0.52}$ & $\mathbf{0.65}$ & $1.66$ & {$+26.8\%$} & {$\downarrow 60.0\%$} \\
\quad + CN Comment + Replan (6) & $0.51$ & $0.62$ & $1.61$ & $+24.4\%$ & {$\downarrow 61.2\%$} \\
\bottomrule
\end{tabularx}
\end{table}

\section{Related Work}
\label{sec:related}

\textbf{Structured Reasoning in LLMs.}
Chain-of-thought (CoT) prompting enables complex reasoning in LLMs\citep{wei2022cot}, but its verbosity and error-proneness have led to more structured alternatives such as Tree-of-Thought (ToT)\citep{yao2023treeofthoughts} and Algorithm-of-Thought (AoT)~\citep{zhang2023algorithm}. Despite improvements in reasoning paths, these methods still rely on natural language, limiting token efficiency and verifiability.

\textbf{Code-Based Reasoning and Token Efficiency.}
Code-enhanced reasoning offers structure and efficiency. Program-Aided LMs (PAL) use Python interpreters~\citep{gao2023pal}, while CodeAct~\citep{wang2024codeact} maps agent actions to executable code. Other efforts~\citep{zhou2023math} adopt code interpreters for math problems. Pseudocode-based approaches like ProgPrompt~\citep{singh2023progprompt} and CodePlan~\citep{wen2025codeplan} yield concise, readable plans. Recent methods further boost token efficiency: Token-Aware Coding Flow reduces token usage by up to 50\% through refactoring~\citep{hu2025tokenaware}, and THINK-AND-EXECUTE improves reasoning via pseudocode simulation~\citep{li2024thinkexecute}. Our method differs by generating structured pseudocode with embedded lightweight feedback at prompt time—eliminating the need for refactoring or execution.

\textbf{Feedback and Replanning Mechanisms.}
Feedback loops improve LLM adaptability. BrainBody-LLM~\citep{bhat2024brainbody} separates planning and control via dual LLMs, supporting closed-loop feedback. Reflexion~\citep{shinn2023reflexion} and Inner Monologue~\citep{huang2023inner} use self-reflection and internal dialogue, though all rely on natural language for feedback. Structured code-based feedback remains underexplored.

\textbf{Multi-Agent Collaboration and Communication.}
Multi-agent frameworks like MegaAgent~\citep{wang2024megaagent} and TalkHier~\citep{wang2025talkhier} enable scalable collaboration via structured protocols. Others—CAMEL~\citep{li2023camel}, MetaGPT~\citep{hong2024metagpt}, AutoGen~\citep{wu2023autogen}, and SmolAgents~\citep{smolagents2024}—employ role-playing and standard workflows. However, most rely on natural language, which increases token costs and introduces ambiguity. Structured code-based communication across agents remains largely untapped.

\textbf{Positioning Our Work.}
We unify pseudocode prompting, feedback, and replanning within a multi-agent framework. Unlike prior work focused on either code or natural language, we adopt a hybrid format that combines structured pseudocode with natural language comments, including bilingual annotations, to improve clarity and flexibility. Ablation studies confirm the benefits of each module. We also extend planning from single-agent to multi-agent settings, highlighting the role of communication protocols and task decomposition in complex scenarios. Our approach is motivated by the limitations of language-only agents—such as ambiguity, weak executability, and inconsistent reasoning—which led us to explore a codified prompting paradigm. By making reasoning explicit and modular, structured prompts offer a more reliable and interpretable interface. We believe this direction moves toward verifiable, tool-integrated AI systems beyond free-form generation.

\section{Conclusion}
This paper presents a codified prompting framework that enhances LLM reasoning by representing agent interactions as typed pseudocode with modular control flows. This structure improves transparency, execution reliability, and token efficiency. Empirical results on GAIA, HotpotQA, and VirtualHome show that our approach consistently outperforms natural language prompting in accuracy, adaptability, and cost-effectiveness, demonstrating its potential for reliable and efficient multi-agent coordination.While our findings are encouraging, the framework's applicability to more diverse, real-world tasks remains to be further explored. Our current experiments focus on representative benchmarks, and results on additional tasks will be released on our GitHub page.Beyond immediate improvements in performance and cost, we believe codified prompting offers a path toward more interpretable and verifiable AI. Looking ahead, integrating multiple language modalities—such as mathematics for formal reasoning, natural language for interpretability, and database querying for information access—with code as the execution layer may unlock more flexible and robust agent capabilities. We hope this work contributes to a broader shift toward structured, tool-integrated, and human-aligned AI systems.

\bibliographystyle{plainnat}
\bibliography{main}

\begin{thebibliography}{24}
\providecommand{\natexlab}[1]{#1}
\providecommand{\url}[1]{\texttt{#1}}
\expandafter\ifx\csname urlstyle\endcsname\relax
  \providecommand{\doi}[1]{doi: #1}\else
  \providecommand{\doi}{doi: \begingroup \urlstyle{rm}\Url}\fi

\bibitem[Bhat et~al.(2024)Bhat, Kaypak, Krishnamurthy, Karri, and Khorrami]{bhat2024brainbody}
Vineet Bhat, Ali~Umut Kaypak, Prashanth Krishnamurthy, Ramesh Karri, and Farshad Khorrami.
\newblock Grounding llms for robot task planning using closed-loop state feedback.
\newblock \emph{arXiv preprint arXiv:2402.08546}, 2024.
\newblock URL \url{https://arxiv.org/abs/2402.08546}.

\bibitem[Gao et~al.(2023)Gao, Madaan, Zhou, Alon, Liu, Yang, Callan, and Neubig]{gao2023pal}
Luyu Gao, Aman Madaan, Shuyan Zhou, Uri Alon, Pengfei Liu, Yiming Yang, Jamie Callan, and Graham Neubig.
\newblock Pal: Program-aided language models.
\newblock \emph{arXiv preprint arXiv:2211.10435}, 2023.
\newblock URL \url{https://arxiv.org/abs/2211.10435}.

\bibitem[Hong et~al.(2024)Hong, Zhuge, Chen, Zheng, Cheng, Wang, Zhang, Yau, Lin, Zhou, Ran, Xiao, Wu, and Schmidhuber]{hong2024metagpt}
Shixiang Hong, Ming Zhuge, Jie Chen, Xue Zheng, Yuxuan Cheng, Jing Wang, Cheng Zhang, Stephen Yau, Zihan Lin, Lei Zhou, Chunlin Ran, Li~Xiao, Chun Wu, and J{\"u}rgen Schmidhuber.
\newblock Metagpt: Meta programming for a multi-agent collaborative framework.
\newblock \emph{arXiv preprint arXiv:2308.00352}, 2024.
\newblock URL \url{https://arxiv.org/abs/2308.00352}.

\bibitem[Hu et~al.(2025)Hu, Zheng, Liu, and Liu]{hu2025tokenaware}
Junwei Hu, Weicheng Zheng, Yan Liu, and Yihan Liu.
\newblock Token-aware coding flow: A study with nano surge in reasoning model.
\newblock \emph{arXiv preprint arXiv:2504.15989}, 2025.
\newblock URL \url{https://arxiv.org/abs/2504.15989}.

\bibitem[Huang et~al.(2023)Huang, Singh, Blukis, and Garg]{huang2023inner}
Wenlong Huang, Ishika Singh, Valts Blukis, and Animesh Garg.
\newblock Inner monologue: Embodied reasoning through planning with language models.
\newblock \emph{arXiv preprint arXiv:2207.05608}, 2023.
\newblock URL \url{https://arxiv.org/abs/2207.05608}.

\bibitem[Li et~al.(2023)Li, Hammoud, Itani, Khizbullin, and Ghanem]{li2023camel}
Guohao Li, Hasan Abed Al~Kader Hammoud, Hani Itani, Dmitrii Khizbullin, and Bernard Ghanem.
\newblock Camel: Communicative agents for "mind" exploration of large language model society.
\newblock \emph{arXiv preprint arXiv:2303.17760}, 2023.
\newblock URL \url{https://arxiv.org/abs/2303.17760}.

\bibitem[Li et~al.(2024)Li, Wang, Lu, et~al.]{li2024thinkexecute}
Yichong Li, Kai Wang, Zhi Lu, et~al.
\newblock Language models as compilers: Simulating pseudocode execution improves algorithmic reasoning in language models.
\newblock \emph{arXiv preprint arXiv:2404.02575}, 2024.
\newblock URL \url{https://arxiv.org/abs/2404.02575}.

\bibitem[Mialon et~al.(2023)Mialon, Fourrier, Swift, Wolf, LeCun, and Scialom]{mialon2023gaia}
Grégoire Mialon, Clémentine Fourrier, Craig Swift, Thomas Wolf, Yann LeCun, and Thomas Scialom.
\newblock Gaia: A benchmark for general ai assistants.
\newblock \emph{arXiv preprint arXiv:2311.12983}, 2023.
\newblock URL \url{https://arxiv.org/abs/2311.12983}.

\bibitem[Puig et~al.(2018)Puig, Ra, Boben, Li, Wang, Fidler, and Torralba]{puig2018virtualhome}
Xavier Puig, Kevin Ra, Marko Boben, Jiaman Li, Tingwu Wang, Sanja Fidler, and Antonio Torralba.
\newblock Virtualhome: Simulating household activities via programs.
\newblock In \emph{Proceedings of the IEEE Conference on Computer Vision and Pattern Recognition (CVPR)}, pages 8494--8502, 2018.
\newblock URL \url{https://openaccess.thecvf.com/content_cvpr_2018/html/Puig_VirtualHome_Simulating_Household_CVPR_2018_paper.html}.

\bibitem[Roucher et~al.(2024)Roucher, Villanova~del Moral, Wolf, von Werra, and Kaunism{\"a}ki]{smolagents2024}
Aymeric Roucher, Albert Villanova~del Moral, Thomas Wolf, Leandro von Werra, and Erik Kaunism{\"a}ki.
\newblock smolagents: a smol library to build great agentic systems.
\newblock \url{https://github.com/huggingface/smolagents}, 2024.
\newblock Accessed: 2025-05-11.

\bibitem[Shinn et~al.(2023)Shinn, Cassano, Gopinath, Narasimhan, and Yao]{shinn2023reflexion}
Noa Shinn, Francesco Cassano, Avinash Gopinath, Karthik~R Narasimhan, and Shinn Yao.
\newblock Reflexion: Language agents with verbal reinforcement learning.
\newblock \emph{arXiv preprint arXiv:2303.11366}, 2023.
\newblock URL \url{https://arxiv.org/abs/2303.11366}.

\bibitem[Singh et~al.(2023)Singh, Blukis, Mousavian, Goyal, Xu, and Tremblay]{singh2023progprompt}
Ishika Singh, Valts Blukis, Arsalan Mousavian, Ayush Goyal, Danfei Xu, and Jonathan Tremblay.
\newblock Progprompt: Generating situated robot task plans using large language models.
\newblock \emph{arXiv preprint arXiv:2209.11302}, 2023.
\newblock URL \url{https://arxiv.org/abs/2209.11302}.

\bibitem[Wang et~al.(2023)Wang, Xu, Lan, Hu, Lan, Lee, and Lim]{wang2023planandsolve}
Liang Wang, Wenxuan Xu, Yanyan Lan, Zhiyuan Hu, Yiming Lan, Raymond Lee, and Ee-Peng Lim.
\newblock Plan-and-solve prompting: Improving zero-shot chain-of-thought reasoning by large language models.
\newblock \emph{arXiv preprint arXiv:2305.04091}, 2023.
\newblock URL \url{https://arxiv.org/abs/2305.04091}.

\bibitem[Wang et~al.(2024{\natexlab{a}})Wang, Wang, Li, Liang, and He]{wang2024megaagent}
Qian Wang, Tianyu Wang, Qinbin Li, Jingsheng Liang, and Bingsheng He.
\newblock Megaagent: A practical framework for autonomous cooperation in large-scale llm agent systems.
\newblock \emph{arXiv preprint arXiv:2408.09955}, 2024{\natexlab{a}}.
\newblock URL \url{https://arxiv.org/abs/2408.09955}.

\bibitem[Wang et~al.(2024{\natexlab{b}})Wang, Chen, Yuan, Zhang, Li, Peng, and Ji]{wang2024codeact}
Xingyao Wang, Yangyi Chen, Lifan Yuan, Yizhe Zhang, Yunzhu Li, Hao Peng, and Heng Ji.
\newblock Executable code actions elicit better llm agents.
\newblock In \emph{Proceedings of the 41st International Conference on Machine Learning (ICML)}, 2024{\natexlab{b}}.
\newblock URL \url{https://arxiv.org/abs/2402.01030}.

\bibitem[Wang et~al.(2025)Wang, Moriyama, Wang, Gangopadhyay, and Takamatsu]{wang2025talkhier}
Zhao Wang, Sota Moriyama, Wei-Yao Wang, Briti Gangopadhyay, and Shingo Takamatsu.
\newblock Talk structurally, act hierarchically: A collaborative framework for llm multi-agent systems.
\newblock \emph{arXiv preprint arXiv:2502.11098}, 2025.
\newblock URL \url{https://arxiv.org/abs/2502.11098}.

\bibitem[Wei et~al.(2022)Wei, Wang, Schuurmans, Bosma, Ichter, Xia, Chi, Le, and Zhou]{wei2022cot}
Jason Wei, Xuezhi Wang, Dale Schuurmans, Maarten Bosma, Brian Ichter, Fei Xia, Ed~Chi, Quoc Le, and Denny Zhou.
\newblock Chain-of-thought prompting elicits reasoning in large language models.
\newblock \emph{arXiv preprint arXiv:2201.11903}, 2022.
\newblock URL \url{https://arxiv.org/abs/2201.11903}.

\bibitem[Wen et~al.(2025)Wen, Guan, Wang, Wu, and Huang]{wen2025codeplan}
Jiarui Wen, Jiahua Guan, Haoyu Wang, Wenhao Wu, and Minlie Huang.
\newblock Codeplan: Unlocking reasoning potential in large language models by scaling code-form planning.
\newblock \emph{arXiv preprint arXiv:2409.12452}, 2025.
\newblock URL \url{https://arxiv.org/abs/2409.12452}.

\bibitem[Wu et~al.(2023)Wu, Bansal, Zhang, Wu, Li, Zhu, Jiang, Zhang, Zhang, Liu, et~al.]{wu2023autogen}
Qingyun Wu, Gagan Bansal, Jieyu Zhang, Yiran Wu, Beibin Li, Erkang Zhu, Li~Jiang, Xiaoyun Zhang, Shaokun Zhang, Jiale Liu, et~al.
\newblock Autogen: Enabling next-gen llm applications via multi-agent conversation.
\newblock \emph{arXiv preprint arXiv:2308.08155}, 2023.
\newblock URL \url{https://arxiv.org/abs/2308.08155}.

\bibitem[Yang et~al.(2018)Yang, Qi, Zhang, Bengio, Cohen, Salakhutdinov, and Manning]{yang2018hotpotqa}
Zhilin Yang, Peng Qi, Saizheng Zhang, Yoshua Bengio, William~W. Cohen, Ruslan Salakhutdinov, and Christopher~D. Manning.
\newblock Hotpotqa: A dataset for diverse, explainable multi-hop question answering.
\newblock In \emph{Proceedings of the 2018 Conference on Empirical Methods in Natural Language Processing}, pages 2369--2380. Association for Computational Linguistics, 2018.
\newblock URL \url{https://aclanthology.org/D18-1259/}.

\bibitem[Yao et~al.(2022)Yao, Zhao, Yu, Zhao, Zhang, and Zhao]{yao2022react}
Shinn Yao, Jeffrey Zhao, Dian Yu, Yanfei Zhao, Yi~Zhang, and Dongyan Zhao.
\newblock React: Synergizing reasoning and acting in language models.
\newblock \emph{arXiv preprint arXiv:2210.03629}, 2022.
\newblock URL \url{https://arxiv.org/abs/2210.03629}.

\bibitem[Yao et~al.(2023)Yao, Zhao, Yu, Zhao, Zhang, and Zhao]{yao2023treeofthoughts}
Shinn Yao, Jeffrey Zhao, Dian Yu, Yanfei Zhao, Yi~Zhang, and Dongyan Zhao.
\newblock Tree of thoughts: Deliberate problem solving with large language models.
\newblock \emph{arXiv preprint arXiv:2305.10601}, 2023.
\newblock URL \url{https://arxiv.org/abs/2305.10601}.

\bibitem[Zhang et~al.(2023)Zhang, Zhang, Li, and Smola]{zhang2023algorithm}
Zhuosheng Zhang, Aston Zhang, Mu~Li, and Alexander Smola.
\newblock Algorithm of thoughts: Enhancing exploration of ideas in large language models.
\newblock \emph{arXiv preprint arXiv:2308.10379}, 2023.
\newblock URL \url{https://arxiv.org/abs/2308.10379}.

\bibitem[Zhou et~al.(2023)Zhou, Wang, Lu, Shi, Luo, Qin, Lu, Jia, Liu, Zhang, et~al.]{zhou2023math}
Aojun Zhou, Kai Wang, Zhi Lu, Weijia Shi, Shuhan Luo, Zhaohui Qin, Shuchang Lu, An~Jia, Yifan Liu, Yichong Zhang, et~al.
\newblock Solving challenging math word problems using gpt-4 code interpreter with code-based self-verification.
\newblock \emph{arXiv preprint arXiv:2308.07921}, 2023.
\newblock URL \url{https://arxiv.org/abs/2308.07921}.

\end{thebibliography}

\end{document}